\documentclass[sigconf,natbib=true]{acmart}

\usepackage{amsmath}

\usepackage{amssymb}
\usepackage{booktabs}
\usepackage{graphicx}
\usepackage{multirow}
\usepackage{xcolor}
\usepackage{algorithm}
\usepackage{algorithmic}
\usepackage{tikz}
\usetikzlibrary{patterns}
\usepackage{pgfplots}
\pgfplotsset{compat=1.18}

\copyrightyear{2026}
\acmYear{2026}
\setcopyright{cc}
\setcctype{by}
\acmConference[CIKM '26]{Proceedings of the 35th ACM International Conference on Information and Knowledge Management}{November 07--11, 2026}{Rome, Italy}
\acmBooktitle{Proceedings of the 35th ACM International Conference on Information and Knowledge Management (CIKM '26), November 07--11, 2026, Rome, Italy}
\acmDOI{10.1145/3799682.3840051}
\acmISBN{979-8-4007-2539-5/2026/11}
\begin{document}

\title{Beyond Uncertainty: Multi-Solver Disagreement Rewards for\\Self-Evolving Reasoning Curricula}

\author{Vinoth Selvendran}
\orcid{0009-0006-6051-1656}
\affiliation{%
  \institution{Independent Researcher}
  \city{Palo Alto}
  \state{CA}
  \country{USA}
}
\email{vinoth459005@gmail.com}

\author{Zhanming Zhang}
\orcid{0009-0006-4965-7558}
\affiliation{%
  \institution{Independent Researcher}
  \city{New York}
  \state{NY}
  \country{USA}
}
\email{zmzhang15@gmail.com}



\begin{abstract}
Self-evolving reasoning frameworks train a Challenger to generate questions exposing a Solver's weaknesses, creating adaptive curricula without human data. However, existing approaches use a single solver's sampling uncertainty as the Challenger's reward. This creates a fundamental bottleneck: as the solver grows confident on the Challenger's question distribution, all sampled answers converge identically, collapsing the reward to zero and starving the Challenger of learning signal. Critically, this single-model reward cannot distinguish genuinely easy questions from those that merely align with one solver's learned biases. We propose a multi-solver disagreement reward using a heterogeneous ensemble varying in model capacity and sampling temperature. A normalized Shannon entropy over the ensemble's per-question plurality answers explicitly rewards questions where solvers produce conflicting solutions---capturing difficulty as inter-model divergence rather than intra-model sampling variance. This richer gradient enables the Challenger to discover questions targeting true capability boundaries, producing a curriculum that forces downstream Solvers to develop robust reasoning strategies generalizing across problem types. Our approach is a drop-in reward function replacement requiring no framework modifications or additional data. Experiments with Qwen3-4B show that Solvers trained on disagreement-Challenger questions achieve +1.34 points average improvement on competition-math benchmarks (MATH-500, AMC, Olympiad), suggesting that multi-solver disagreement provides a complementary and scalable signal for curriculum generation in self-play reasoning systems.
\end{abstract}

\maketitle

\section{Introduction}
\label{sec:intro}

Large language models (LLMs) have achieved remarkable reasoning capabilities through reinforcement learning from human feedback~\cite{rlhf} and supervised fine-tuning~\cite{grpo}. However, curating high-quality training data at scale remains expensive. Data-free self-evolution~\cite{rzero, azr} addresses this by employing a \emph{Challenger--Solver} co-evolutionary loop: the Challenger generates problems targeting the Solver's capability frontier, while the Solver improves by solving increasingly difficult Challenger-posed problems.

R-Zero~\cite{rzero} operationalizes this paradigm by rewarding the Challenger proportionally to the Solver's sampling uncertainty---measured as inconsistency across $m$ rollouts. When the Solver answers correctly approximately half the time ($\hat{p} \approx 0.5$), the question sits at its capability edge, yielding maximum reward. This uncertainty-driven curriculum produces substantial gains: +6.49 on math and +7.54 on general reasoning from Qwen3-4B-Base.

However, single-solver uncertainty conflates two distinct phenomena. A question may appear ``easy'' (high $\hat{p}$, low reward) because it aligns with one model's learned biases, while being genuinely challenging for a differently-trained model. R-Zero reports that pseudo-label accuracy declines progressively from 79\% to 63\% across three iterations (Section~4.4, Table~5 of~\cite{rzero}), with model performance degrading after iteration~3---evidence that single-model uncertainty eventually becomes an unreliable proxy for genuine difficulty.

In this work, we augment the Challenger reward with a \textbf{multi-solver disagreement signal} that measures inter-model divergence across a heterogeneous ensemble. Our approach computes normalized Shannon entropy over per-variant plurality answers, explicitly rewarding questions where solvers produce conflicting solutions. The method is implemented as a drop-in reward function replacement requiring no modifications to the training framework.\footnote{All reported deltas ($\Delta$) are absolute accuracy points throughout this paper.}

Our contributions are as follows: (1)~We propose a normalized Shannon entropy reward over a 3-solver ensemble that provides a signal trending upward (0.61 to 0.72) within a single training iteration, while uncertainty reward remains stable at ${\approx}\,0.20$. (2)~We show that the disagreement reward redistributes curriculum toward an intermediate-difficulty band: +1.34 points on competition-math (MATH-500/AMC/Olympiad), with a minor regression of $-$0.60 points on GSM8K due to reduced curriculum coverage of routine problems. (3)~The approach requires zero modifications to the RL training framework or data pipeline.

\begin{figure*}[t]
\centering
\includegraphics[width=\textwidth]{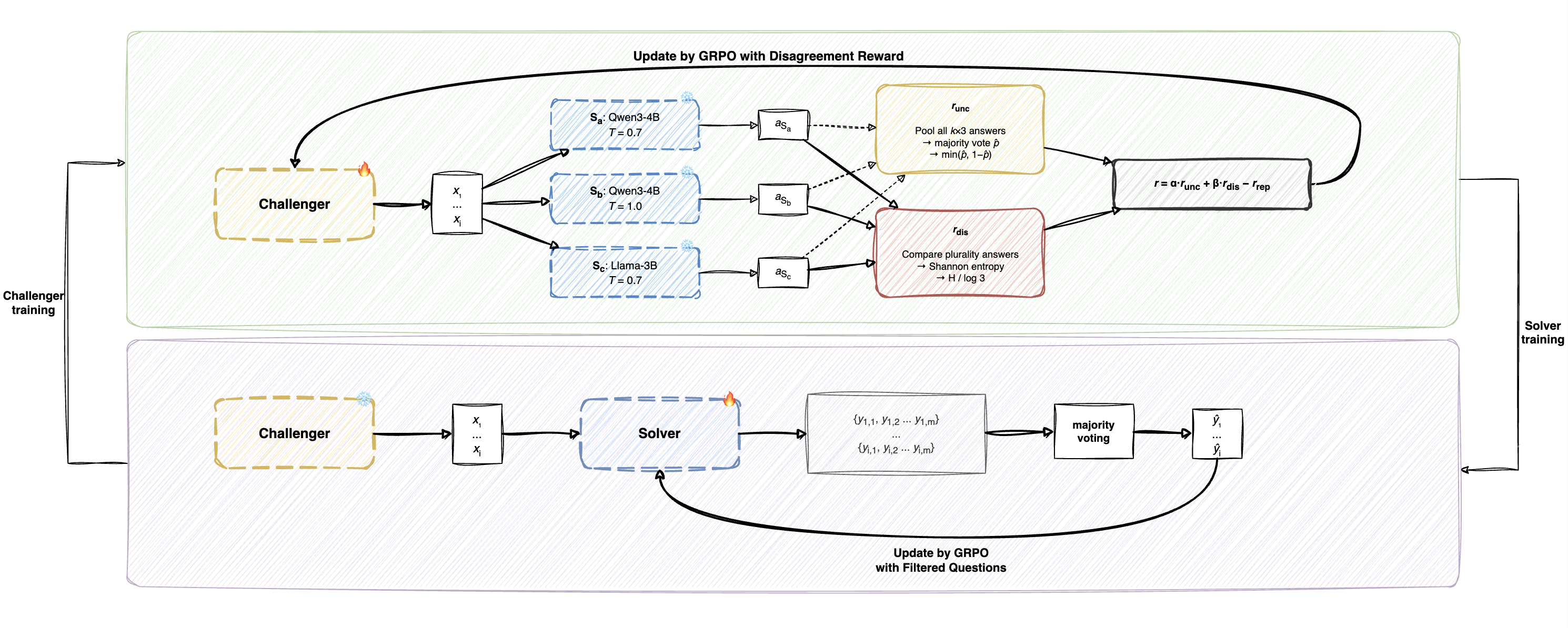}
\caption{An overview of our framework, which extends the R-Zero~\cite{rzero} Challenger--Solver co-evolution architecture. \textbf{Top (Challenger training):} The Challenger generates questions that are dispatched to a heterogeneous ensemble of three frozen solver variants. Each variant independently produces a plurality answer; these feed into two separate reward paths---$r_\text{unc}$ (pooled uncertainty over all samples) and $r_\text{dis}$ (cross-variant disagreement via normalized Shannon entropy). The combined reward drives GRPO updates to the Challenger policy. \textbf{Bottom (Solver training):} The Solver is trained on filtered questions from the disagreement-Challenger using majority-voted pseudo-labels.}
\label{fig:overview}
\end{figure*}

\section{Related Work}
\label{sec:related}

\textbf{Data-free self-evolution and reward shaping.} Self-play for LLM reasoning, inspired by game-playing agents~\cite{self_play_silver}, has produced several data-free training frameworks. R-Zero~\cite{rzero} introduced the Challenger--Solver co-evolution loop where a question-generator is rewarded by the solver's sampling uncertainty, producing adaptive curricula without any labeled data. Absolute Zero Reasoner~\cite{azr} simplifies this to single-model self-play using code executors as verifiers, while Socratic-Zero~\cite{socratic} extends to three-agent co-evolution with specialized roles. STaR~\cite{star} bootstraps reasoning by iteratively training on self-generated rationales that lead to correct answers, and ReST~\cite{rest} generalizes this with quality-filtered self-generated data. A shared vulnerability across these approaches is reward collapse: Co-rewarding~\cite{stable_self_rl} identifies ``self-consistent illusions'' where single-model supervision converges to confidently wrong solutions that the model itself cannot detect, and Pan et al.~\cite{reward_hacking} demonstrate that this failure mode compounds over multiple iterations of self-refinement. RLHF~\cite{rlhf} mitigates this by grounding rewards in human preferences, but at substantial annotation cost. Our work introduces cross-model grounding as an alternative that requires no human annotation.

\textbf{Ensemble disagreement for data selection.} Using ensemble disagreement to identify informative samples has a long history. Query-by-committee~\cite{query_by_committee} selects samples where committee members maximally disagree, and Dietterich~\cite{ensemble_diversity} provides theoretical foundations for why ensemble diversity improves generalization. Deep active learning~\cite{active_learning} extends these principles to neural networks, while semi-supervised methods~\cite{semi_supervised} use ensemble confidence to guide pseudo-label selection. Curriculum learning~\cite{curriculum_learning} and self-paced learning~\cite{self_paced} order training data by difficulty, typically using model confidence as a proxy. Our work departs from these approaches by applying disagreement not to data filtering from a fixed pool, but to the \emph{reward signal} of a generative RL curriculum---steering the data generator itself toward producing harder questions.

\textbf{Mathematical reasoning.} DeepSeek-R1~\cite{deepseek_r1} demonstrates that RL alone can elicit sophisticated reasoning behaviors including self-verification. GRPO~\cite{grpo} provides efficient policy optimization for reasoning tasks by computing advantages relative to group samples. Process reward models~\cite{prm} provide step-level supervision for multi-step problem solving. We evaluate on standard benchmarks including MATH~\cite{math_benchmark} and GSM8K~\cite{gsm8k}, using Qwen3~\cite{qwen3} and Llama~\cite{llama3} as foundation models.

\section{Notations and Preliminaries}
\label{sec:prelim}

Let $Q_\theta$ denote the Challenger model parameterized by $\theta$, and let $\mathcal{E} = \{S_1, \ldots, S_N\}$ denote an ensemble of $N$ solver variants. For a generated question $q$, each variant $S_i$ produces $k$ sampled answers $\{y_1^{(i)}, \ldots, y_k^{(i)}\}$. Let $a_i$ denote the intra-variant plurality answer for $S_i$, computed via majority vote with symbolic equivalence matching. Let $\hat{p}$ denote the empirical accuracy of the majority-vote pseudo-label $\tilde{y}$ over the pooled answer set, and $H(\cdot)$ denote Shannon entropy in natural logarithms. The composite Challenger reward is defined as $r(q) = \alpha \cdot r_\text{unc}(q) + \beta \cdot r_\text{dis}(q) - r_\text{rep}(q)$, where $r_\text{unc}$, $r_\text{dis}$, and $r_\text{rep}$ denote the uncertainty, disagreement, and repetition penalty terms respectively.

In the R-Zero framework~\cite{rzero}, the Challenger and Solver are trained alternately via Group Relative Policy Optimization (GRPO)~\cite{grpo}. During Challenger training, the Solver is frozen; during Solver training, the Challenger is frozen. This alternating optimization produces a co-evolutionary curriculum where question difficulty and solver capability advance together.

\section{Method}
\label{sec:method}

\subsection{Uncertainty Reward (Baseline)}

In R-Zero, the Solver $S_\phi$ produces $m$ answers for each question $q$. The pseudo-label $\tilde{y}$ is determined by majority vote with empirical accuracy:
\begin{equation}
    \hat{p} = \frac{1}{m}\sum_{j=1}^{m} \mathbf{1}[y_j = \tilde{y}]
\end{equation}
The uncertainty reward is defined as:
\begin{equation}
    r_\text{unc}(q) = \min(\hat{p},\; 1{-}\hat{p})
\end{equation}
which peaks at $\hat{p} = 0.5$ and yields zero signal when the Solver answers consistently. A BLEU-based repetition penalty $r_\text{rep}$ is computed via agglomerative clustering ($\tau{=}0.5$, average linkage) over generated questions within each batch.

The limitation of $r_\text{unc}$ is that it is bounded by the variance of a single model's sampling distribution. As the Solver grows confident on the Challenger's output distribution, $\hat{p} \to 1$ and $r_\text{unc} \to 0$, regardless of whether the question is genuinely easy or merely aligned with that model's biases.

\subsection{Ensemble Disagreement Reward}

We introduce a heterogeneous ensemble $\mathcal{E} = \{S_a, S_b, S_c\}$ consisting of three solver variants:
\begin{itemize}
    \item $S_a$: Solver v1 checkpoint (Qwen3-4B, GRPO-trained), $T{=}0.7$
    \item $S_b$: Same Solver v1 weights, $T{=}1.0$ (sampling diversity)
    \item $S_c$: Llama-3.2-3B-Instruct, $T{=}0.7$ (architectural diversity)
\end{itemize}

The ensemble varies along two axes: sampling temperature ($S_a$ vs.\ $S_b$) and model architecture ($S_a$ vs.\ $S_c$). This design captures both intra-architecture sensitivity and inter-architecture divergence.

For each question $q$, each variant $S_i$ independently generates $k{=}10$ samples. The intra-variant plurality answer $a_i$ is computed as follows. First, all $k$ answers are clustered via pairwise symbolic math equivalence checking using \texttt{mathruler.grade\_answer}, which performs symbolic simplification and numerical comparison with tolerance $\epsilon{=}10^{-6}$. Equivalent answers are merged via a Union-Find data structure. The representative of the largest cluster is selected as $a_i$.

The disagreement reward is defined as the normalized Shannon entropy:
\begin{equation}
    r_\text{dis}(q) = \frac{H(a_{S_a}, a_{S_b}, a_{S_c})}{\log |\mathcal{E}|}
\end{equation}
where $H$ is computed over the distribution of answer-equivalence classes among the plurality answers. Normalization by $\log|\mathcal{E}| = \log 3$ bounds the reward to $[0,1]$: unanimous agreement yields $r_\text{dis} = 0$; three mutually distinct answers yield $r_\text{dis} = 1$; a 2-vs-1 split yields $r_\text{dis} \approx 0.58$. When fewer than 2 variants respond (due to timeout), the implementation normalizes by $\log(N_\text{responding})$ rather than $\log|\mathcal{E}|$.

\subsection{Composite Reward}

The composite Challenger reward combines the two signals:
\begin{equation}
    r(q) = \alpha \cdot r_\text{unc}(q) + \beta \cdot r_\text{dis}(q) - r_\text{rep}(q)
\end{equation}
with $\alpha{=}1.0$, $\beta{=}0.5$. Here $r_\text{unc}$ is computed over the pooled bag of all $3k{=}30$ samples, preserving upstream semantics, while $r_\text{dis}$ captures cross-variant divergence. The asymmetric weighting ($\alpha > \beta$) preserves uncertainty as the primary signal while using disagreement as a complementary term. Under default weights, the analytical bounds are $r(q) \in [-1, 1]$.

Parse failures (malformed outputs without valid \texttt{$\backslash$boxed\{\}} content) receive $r(q) = -1$ as a format penalty, matching R-Zero semantics.

\subsection{Complementarity of $r_\text{unc}$ and $r_\text{dis}$}

The two terms capture orthogonal information. $r_\text{unc}$ measures collective sampling noise---it peaks when the pooled answer distribution is roughly split. $r_\text{dis}$ measures inter-model divergence---questions where differently-trained models reach different conclusions even when individually confident. A question can have low $r_\text{unc}$ but high $r_\text{dis}$: if $S_a$ and $S_b$ each produce 10/10 answers for ``$A$'' while $S_c$ produces 10/10 for ``$B$'', the pooled bag gives $\hat{p} = 0.67$, $r_\text{unc} = 0.33$, but $r_\text{dis} = 0.58$ (2-vs-1 split).

\subsection{System Design}

We design a distributed reward computation system consisting of three dedicated inference engines (one per solver variant), each on a separate GPU with a lightweight HTTP interface. Algorithm~\ref{alg:reward} summarizes the per-question computation. During Challenger training, each rollout is parsed to extract the generated question and reference answer. Valid questions are dispatched to all three engines concurrently via a thread pool executor with a 30-second per-engine timeout. Each engine performs $k{=}10$ independent generations, returning raw answer strings for Union-Find clustering and reward computation.

The system handles edge cases gracefully: (i)~variant timeout triggers degradation to responding variants only; (ii)~parse failures receive a fixed penalty; (iii)~batch-level BLEU clustering operates over successfully parsed questions only. The fan-out architecture adds ${\sim}8$ minutes wall-clock per batch compared to single-solver evaluation, with ${\sim}3{\times}$ total GPU-hour overhead.

\begin{algorithm}[t]
\caption{Ensemble Disagreement Reward Computation}
\label{alg:reward}
\small
\begin{algorithmic}[1]
\REQUIRE Generated question $q$, ensemble $\mathcal{E} = \{S_a, S_b, S_c\}$
\ENSURE Composite reward $r(q)$
\STATE Parse $q$ from Challenger output; if parse fails, \textbf{return} $r = -1$
\FOR{each variant $S_i \in \mathcal{E}$ \textbf{in parallel}}
    \STATE Generate $k = 10$ answers $\{y_1^{(i)}, \ldots, y_k^{(i)}\}$
    \STATE Cluster answers via Union-Find over \texttt{grade\_answer}
    \STATE $a_i \leftarrow$ representative of largest equivalence class
\ENDFOR
\STATE Pool all $3k$ answers; compute $\hat{p}$ via largest cluster share
\STATE $r_\text{unc} \leftarrow \min(\hat{p},\; 1 - \hat{p})$
\STATE $r_\text{dis} \leftarrow H(a_{S_a}, a_{S_b}, a_{S_c}) \;/\; \log|\mathcal{E}|$
\STATE $r_\text{rep} \leftarrow$ BLEU cluster share (batch-level)
\STATE \textbf{return} $r = \alpha \cdot r_\text{unc} + \beta \cdot r_\text{dis} - r_\text{rep}$
\end{algorithmic}
\end{algorithm}

\section{Experiments}
\label{sec:experiments}

\subsection{Experimental Setup}

We build on R-Zero using Qwen3-4B-Base~\cite{qwen3} as backbone, with GRPO at official hyperparameters (batch size 16, 6 Questioner steps, 20 Solver steps, $m{=}10$ samples). Hardware consists of 8$\times$NVIDIA H100-80GB: four for the FSDP trainer and three for the ensemble inference servers. $S_a$/$S_b$ share the Solver v1 checkpoint at temperatures 0.7/1.0; $S_c$ is Llama-3.2-3B-Instruct at $T{=}0.7$. The baseline is standard R-Zero iter-1 with single-solver uncertainty reward. Our protocol runs one additional Challenger iteration with the disagreement reward, followed by Solver training on the generated curriculum and evaluation. Total training time is approximately 14 GPU-hours per Challenger iteration.

\subsection{Results}

\begin{table}[t]
\centering
\caption{Solver accuracy (pass@1 \%). All models evaluated using the same pipeline with \texttt{mathruler} for answer grading. Disagreement curriculum redistributes training signal toward competition math (+1.34 pts avg).}
\label{tab:main_results}
\small
\begin{tabular}{l|cccc}
\toprule
\textbf{Benchmark} & \textbf{Base} & \textbf{Post-tr.} & \textbf{R-Zero} & \textbf{Ours} \\
\midrule
MATH-500 & 43.51 & 49.40 & 59.20 & \textbf{60.40} \\
AMC & 36.80 & 39.45 & 44.38 & \textbf{45.70} \\
Olympiad & 19.93 & 20.30 & 29.48 & \textbf{30.96} \\
\midrule
\textit{Comp. Avg} & \textit{33.41} & \textit{36.38} & \textit{44.35} & \textit{\textbf{45.69}} \\
\midrule
GSM8K & 78.32 & 89.76 & \textbf{89.99} & 89.39 \\
\bottomrule
\end{tabular}
\vspace{4pt}
\parbox{\columnwidth}{\scriptsize \textit{Base}: Qwen3-4B-Base. \textit{Post-tr.}: Qwen3-4B (post-trained). \textit{R-Zero}: iter-1 solver trained with single-solver uncertainty reward. \textit{Ours}: solver trained with disagreement-Challenger curriculum.}
\end{table}

Table~\ref{tab:main_results} reports pass@1 accuracy across four benchmarks for all models evaluated under the same pipeline. The disagreement-Challenger curriculum produces consistent improvements on competition-level math: +1.20 on MATH-500, +1.32 on AMC, and +1.48 on Olympiad (average +1.34 points over R-Zero). GSM8K shows a minor regression ($-$0.60 pts) attributable to reduced curriculum size.

\textbf{Note on evaluation.} Our absolute scores differ from those reported in the R-Zero paper due to differences in the answer verification pipeline. We use \texttt{mathruler} with symbolic equivalence checking and Claude Sonnet 4.6 as a recheck judge, whereas the original R-Zero evaluation uses GPT-4o as judge. All comparisons in this work (Base, Post-trained, R-Zero, and Ours) are evaluated under the same pipeline to ensure fair relative comparison; the deltas ($\Delta$) between R-Zero and Ours are the primary metric of interest.

\subsection{Reward Dynamics}

\begin{table}[t]
\centering
\caption{Per-batch reward statistics during Challenger training. $r_\text{dis}$ trends upward (+18\% end-to-end) with decreasing variance; $r_\text{unc}$ remains stable.}
\label{tab:reward_dynamics}
\small
\begin{tabular}{c|cc|cc}
\toprule
\textbf{Batch} & $r_\text{unc}$ & $\sigma$ & $r_\text{dis}$ & $\sigma$ \\
\midrule
1 & .191 & .095 & .609 & .336 \\
2 & .199 & .090 & .639 & .326 \\
3 & .206 & .086 & .637 & .329 \\
4 & .200 & .092 & .619 & .325 \\
5 & .201 & .088 & .636 & .324 \\
6 & .205 & .088 & .671 & .310 \\
7 & .209 & .090 & .701 & .295 \\
8 & .206 & .083 & .718 & .278 \\
\bottomrule
\end{tabular}
\end{table}

Table~\ref{tab:reward_dynamics} reports per-batch statistics measured from training logs. Three observations emerge:

(1)~$r_\text{dis}$ trends upward from 0.609 to 0.718 (+18\% end-to-end) with minor non-monotonic fluctuations in early batches (e.g., 0.637$\to$0.619 at batch 3$\to$4). The standard deviation decreases from 0.336 to 0.278 (17\% reduction), indicating convergence toward a consistent disagreement-rich question distribution.

(2)~$r_\text{unc}$ remains stable at $\approx$0.20 throughout (range 0.191--0.209), suggesting the disagreement signal provides gradient independent of single-solver uncertainty. We note this is observed within a single iteration (8 batches); whether $r_\text{dis}$ avoids multi-iteration collapse~\cite{rzero} is untested.

(3)~$r_\text{rep}$ averages 0.001--0.004, indicating that disagreement naturally promotes question diversity without requiring the BLEU penalty to contribute significantly.

\section{Conclusion}
\label{sec:conclusion}

We proposed a multi-solver disagreement reward for self-evolving reasoning curricula that augments the standard uncertainty signal with normalized Shannon entropy over a heterogeneous solver ensemble. The upward $r_\text{dis}$ trajectory and targeted competition-math gains (+1.34 points on MATH-500/AMC/Olympiad) suggest that inter-model disagreement captures a learnable difficulty dimension complementary to single-solver uncertainty. The approach requires no modifications to the underlying RL framework and is applicable to any Challenger--Solver co-evolution system.

\section*{GenAI Usage Disclosure}
The authors used generative AI in two capacities. First, as part of the evaluation methodology, Claude Sonnet 4.6 (Anthropic) served as an auxiliary recheck judge alongside \texttt{mathruler} symbolic equivalence checking to verify answer correctness, as described in Section~\ref{sec:experiments}. Second, generative AI was used for proofreading and language editing of the manuscript. No generative AI was used for research ideation, experimental design, or data generation. The authors take full responsibility for the content of this paper.

\bibliographystyle{ACM-Reference-Format}
\bibliography{references}

\end{document}